\documentclass[times, twocolumn, final]{elsarticle}
\usepackage[utf8]{inputenc}
\usepackage{amsfonts}
\usepackage{amsmath}
\usepackage{amssymb}
\usepackage{semantic}
\usepackage{subcaption}
\usepackage{graphicx}
\usepackage{xcolor}
\usepackage{wrapfig}
\usepackage{multirow}
\journal{Pattern Recognition Letters}

\begin{document}
\begin{frontmatter}

\title{A Method for Evaluating Deep Generative Models of Images via Assessing the Reproduction of High-order Spatial Context}

\author[1]{Rucha Deshpande}
\author[2]{Mark A. Anastasio}
\author[2]{Frank J. Brooks\corref{cor1}}
\cortext[cor1]{Corresponding author.}
\ead{fjb@illinois.com}

\address[1]{\small Dept. of Biomedical Engineering, Washington University in St. Louis, USA}
\address[2]{Dept. of Bioengineering, University of Illinois Urbana-Champaign, USA}

\begin{abstract}
Deep generative models (DGMs) have the potential to revolutionize diagnostic imaging. Generative adversarial networks (GANs) are one kind of DGM which are widely employed. The overarching problem with deploying GANs, and other DGMs, in any application that requires domain expertise in order to actually use the generated images is that there generally is not adequate or automatic means of assessing the domain-relevant quality of generated images. In this work, we demonstrate several objective tests of images output by two popular GAN architectures. We designed several stochastic context models (SCMs) of distinct image features that can be recovered after generation by a trained GAN. Several of these features are high-order, algorithmic pixel-arrangement rules which are not readily expressed in covariance matrices. We designed and validated statistical classifiers to detect specific effects of the known arrangement rules. We then tested the rates at which two different GANs correctly reproduced the feature context under a variety of training scenarios, and degrees of feature-class similarity. We found that ensembles of generated images can appear largely accurate visually, and show high accuracy in ensemble measures, while not exhibiting the known spatial arrangements. Furthermore, GANs trained on a spectrum of distinct spatial orders did not respect the given prevalence of those orders in the training data. The main conclusion is that SCMs can be engineered to quantify numerous errors \emph{per image} that may not be captured in ensemble statistics but plausibly can affect subsequent use of the GAN-generated images. 
\end{abstract}


\end{frontmatter}%

\section{Introduction}
A deep generative model (DGM) is one where a deep neural network is used to learn to draw variates from an unknown, and typically very high-dimensional, distribution \cite{bond2021deep}. Generative adversarial networks (GANs) \cite{goodfellow2020generative} are one kind of DGM that can be used to draw whole-image variates from an empirical distribution learned from an ensemble of training images. GANs have been designed and deployed in numerous imaging applications such as data augmentation, anomaly detection, and as regularizing priors in image reconstruction problems \cite{kazeminia2020gans, shamsolmoali2021image}. Despite a relatively long history of proposed deployments, objective evaluation of generated images remains an active area of research in imaging and computer vision \cite{borji2021pros}.

When deploying DGMs in any mission-critical application, it is vital to have \emph{objective} measures of image quality \cite{barrett2013foundations} which are well beyond subjectively ``looking good'' to untrained human observers. In applications involving significant domain expertise, such as biomedical or diagnostic imaging, designing objective measures can be especially challenging because it usually is not clear which computable features, if any, express the knowledge of the domain expert. The connection between domain-relevant objective image quality measures and traditional measures of the similarity between feature distributions typically derived from natural images, also remains unclear. Furthermore, because the downstream usage of a generated image likely is nuanced and task-dependent, the very notion of ``ground truth" in a generated image likely is, at best, ambiguous. Therefore, a reasonable starting point toward comprehensive objective assessment of image quality, is to measure the general capacity of a GAN (or other DGM) to reproduce sophisticated contextual features that are known prior to training.  

In this work, we propose the purposeful design of stochastic context models (SCMs) that encode domain-relevant, external knowledge---henceforth, ``spatial context''---and to use the per-image rate of spatial context reproduction as an objective assessment of the capacity of any generative model of images. Here, spatial context may be implicit, i.e., arising from the co-occurrence of image features, or explicit, i.e., an ineluctable pixel-placement rule defined by a human user. Thus, a single generated image will be considered useful if whatever context necessary to perform a downstream task is exactly present. The role of the proposed SCMs is similar to that of stochastic object models (SOMs)---which are commonly employed in the development of imaging systems---in that each serve as a ground truth; however, there is a key difference. Here, the SCMs are generic models of a variety of task-relevant spatial contexts which can appear across a gamut of SOMs and, therefore, should not be thought of as an attempt to model any one particular object or system.

To be clearer still, we do not propose to accurately model any particular object or image for any particular application. Instead, we propose to model some kinds of relative pixel arrangements that are generally important across many applications at once \cite{badano2017much}. For example, in a chest radiograph of a human, there is a known number, location, and size of ``heart features'' \emph{relative to} ``lung features.'' Here, we are not proposing to model hearts or lungs, but, instead, propose to model the frequency and relative location of sophisticated image features. The recoverable spatial context that we propose to encode within each training datum reflects both external and high-order knowledge of correct spatial arrangements. It is external in the practical sense that what should be true about every image cannot be learned from any one image; it is high-order in the sense that correct appearance of features in any one image is not readily expressible in, or detected via, grayscale histograms or variance-covariance matrices. Therefore, we also explicitly note that throughout this work ``order'' should not be confused with the \emph{degree} of moments of any particular probability distribution.

\subsection{Overview of the proposed methodology}
In this work, we demonstrate that both implicit and explicit spatial context can be built into training images algorithmically, such that it can be verified readily after generation, and without specifying formulas for describing any particular image feature. This means that we have a ground truth for testing generated images for various contexts. We then employed  distinct SCMs in several experiments to assess the extent that GANs learned high-order information along with whatever low-order information was learned during training. We have previously reported preliminary results with this approach \cite{deshpande2022evaluating}; the present work provides an additional SCM, mathematical formulations of all SCMs, in addition to updated designs of the original stochastic models as well as rigorous analysis going well beyond the initial work. The goal of this work is to provide a data-driven method, independent of generative model architecture, that enables the assessment of DGMs for their capacity to reproduce domain-relevant, high-order spatial context.

\section{Methods}

\subsection{Description of the SCMs}
\begin{figure}[h!tb]
\centering
\begin{subfigure}{0.45\textwidth}%
    \includegraphics[width=0.9\linewidth]{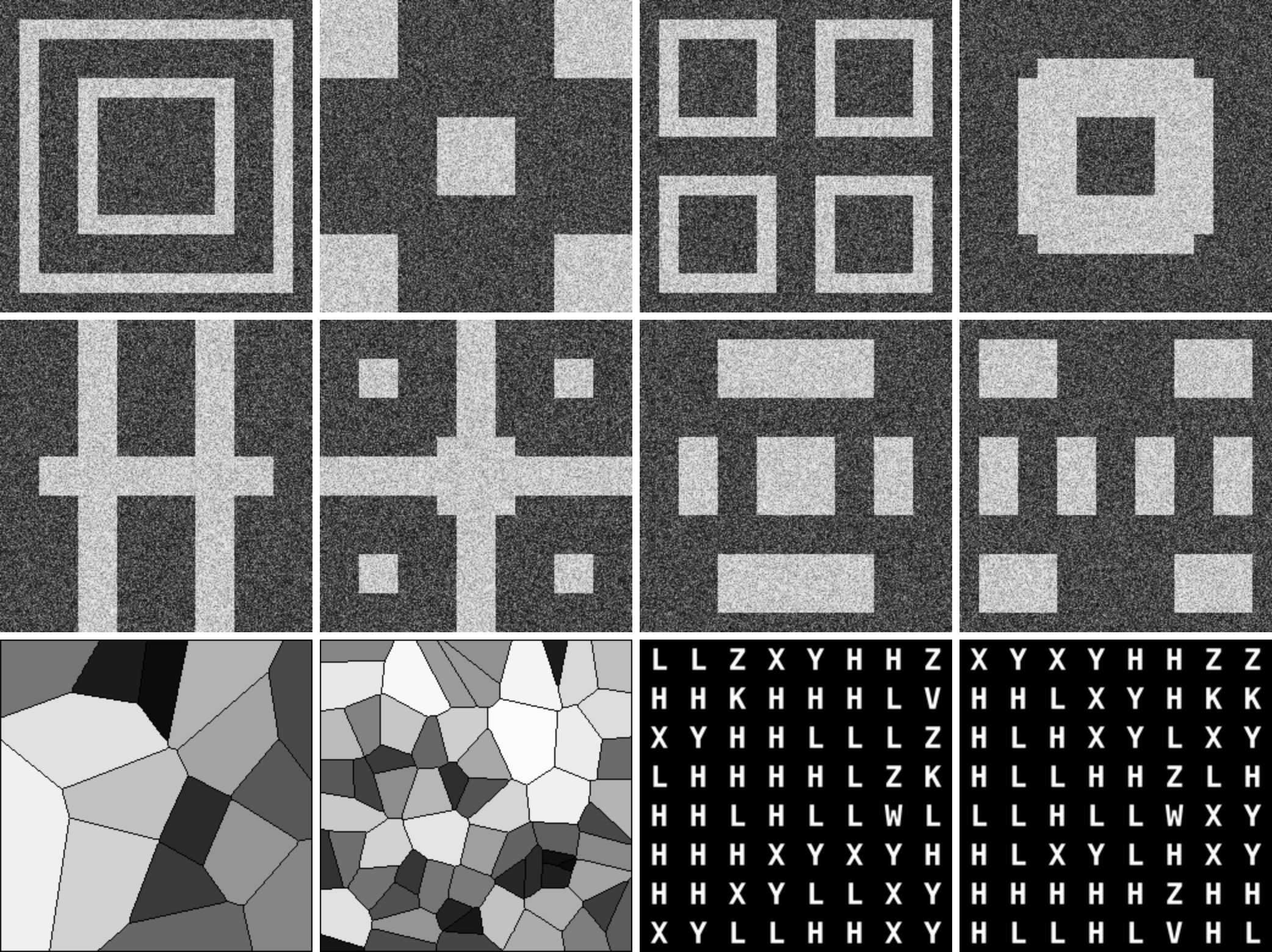} 
\end{subfigure}
\caption{Sample realizations from the three purposefully designed SCMs . Top two rows: One realization each from the eight classes in the flags SCM. Bottom row, left to right: Realizations from the shaded Voronoi SCM representing classes 16 and 64, and the alphabet SCM are shown.}
\label{real_samples}
\end{figure}

Three families of SCMs are described in the following subsections. All realizations from all SCMs are 8-bit grayscale, 256x256-pixel images; sample realizations are shown in Fig.~\ref{real_samples}. The three ensembles (in order of presentation) comprised 32768 (per class), 65536 (per class) and 131072 images respectively. Ensembles from the designed SCMs have been made available on Harvard Dataverse: https://doi.org/10.7910/DVN/HHF4AF. 

\subsubsection{Flags SCM}
We designed the eight-class flags SCM for testing the joint reproducibility of pre-specified, first-, second-, and high-order image features at once. Each image $I$ in any class $c$, can be delineated into a regular grid of 16$\times$16 pixel tiles with each tile corresponding to either foreground $f_k$ or background $b_k$, where $k$ is the tile index, indicating tile location within the grid. Furthermore, $I_c = \{80\times f_k, 176\times b_k\}\:\forall c$; this eliminates the zero-order variance in the number of pixels of interest. 

Any realization in a class can be represented as:
\begin{equation}
\label{flags_main}
    I_c = \sum_k(a_{kc}f_k + (1-a_{kc})b_k), 
\end{equation}
where $\mathcal{A}\in\{0,1\}^{K\times C}$ is a binary matrix indicating background (0) or foreground (1) for all $K$ tile indices in $C$ classes. Thus, $\mathcal{A}$ indicates the prescribed, class-specific foreground patterns, and an image class is one of eight distinct foreground arrangements. 

Grayscale variates within $f_k$ and $b_k$ were chosen from distinct Beta distributions:
\begin{equation}
\label{flags_fg}
    f_k\sim 152\:X + 96,\; \mathrm{where}\; X \sim \mathrm{Beta}(\alpha=4,\beta=2), 
\end{equation}
\begin{equation}
\label{flags_bg}
    b_k\sim 192\:X + 8,\; \mathrm{where}\; X \sim \mathrm{Beta}(\alpha=2,\beta=4). 
\end{equation}

The result was an image with foreground brighter than the background. Moreover, the placement of the variates for $f_k$ and $b_k$ was completely random within $f_k$ and $b_k$. 

Last, a set of certain 24 tile-location indices $k$ was never part of the foreground, in any class:
\begin{equation}
\label{flags_forbidden}
    n = \{k:a_{kc}=0\} \quad \forall c.    
\end{equation}
This is analogous to structural constraints in location for a feature. Together, these classes enable a variety of experiments for exploring how much of each informational order the GAN learns. For example, the extent that learning the correct foreground structures and random arrangements (second-order) also means learning the correct grayscale intensity distributions (first-order) while never misplacing a foreground square in a forbidden location (high-order) can be tested. Furthermore, we can also measure the prevalence of classes in the generated ensemble; class prevalence is one example of external, domain-specific knowledge.

The tiled nature of the images eased post-hoc classification. Each tile in $I$ was identified as $b_k$ or $f_k$ by comparing its intensity mean against a threshold of 140 chosen to be halfway between the two modes of the combined grayscale distribution of $b_k$ and $f_k$. The class ($c$) was then determined by computing the mean absolute error against each column in $\mathcal{A}$. 

\subsubsection{Voronoi SCM}
The Voronoi \cite{boots2009spatial} SCM,  a four-class SCM,  enabled testing of second- and high-order information from randomized sets of image features. Each image $I$ can be represented as a union of the set $V$ of Voronoi regions $v_i$ and their edges $e$. Here, $i=\{1,2,...,c\}$, where $c\in\{16, 32, 48, 64\}$ represents the cardinality of $V$ within each $I$, as well as the image class. Within each $I$, region centers were placed in a spatially random manner, unlike the fixed foreground locations in the flags SCM; this provided an additional source of object variance. Edges $e$ were set to an intensity level of 0; this enabled robust segmentation of $e$ and $v_i$ from a given $I$. All pixels in a $v_i$ were allocated a single grayscale value $g$ drawn from a set of 64 predetermined, equidistant values between 8 and 255. Most importantly, the grayscale value increased monotonically with area, which is a high-order feature: 
\begin{equation}
\label{voronoi}
    \rho(\mathrm{area}(v_i), g) = 1,
\end{equation}
where $\rho$ is Spearman rank-order correlation coefficient. 
 In case of the \textit{unshaded} Voronoi experiment (see Sec.3.2),  all regions $v_i$ were set to a grayscale value of 255. The Voronoi SCM is representative of images with multiple, positionally independent regions of interest within an image, each having a distinct intensity, e.g., histology images. The Voronoi SCM also allowed for testing the ensemble class prevalence, but with feature sets at multiple spatial scales, simultaneously. For the analysis of generated images, post-processing involved identification of the edges $e$, by thresholding each $I$ against an intensity level of 64 for the unshaded Voronoi, or via Sauvola thresholding for the shaded Voronoi, followed by skeletonization. The skeleton was then employed for detecting $v_i$, which in turn determined $c$ and enabled the extraction of region-wise values of $g$. It is noted that although this method of region detection is not perfect, it is still sufficiently robust for the experiments proposed. Calibration of this method on the training data predicted the mean detected number of regions exactly, with standard deviations of 0.1, 0.3, 0.4 and 0.6 for the four classes sequentially. 

\subsubsection{Alphabet SCM}
 Each realization $I$ from this SCM can be delineated into a grid, yielding $32\times32$ pixel tiles $t$ such that each $t$ represents a letter in the alphabet $\mathbb{A} = \{H, K, L, V, W, X, Y, Z\}$. The per-realization prevalence of all letters within the image $I$ was fixed according to the prescribed set $\mathbb{B} = \{24 \times H, 2 \times K, 16 \times L, 1 \times V, 1 \times W, 8 \times X, 8 \times Y, 4 \times Z\}$. Thus, each realization can be represented as:
 \begin{equation}
 \label{alphabet_main}
     I = \{t_{r,c} : \bigcup\limits_{r,c}f(t_{r,c})=\mathbb{B}\},
 \end{equation}
 where $f(t):t^{32\times32}\to\mathbb{A}$ represents a template matching operation, and $r,c$ are respectively the row and column indices of $t$ within the grid. In other words, each image $I$ comprised letter-tiles $t_{r,c}$ that together represent the complete set of specific letters at prescribed prevalences, i.e., $\mathbb{B}$. Although the locations of specific letters within $I$ could vary---thus, providing random variation across realizations---they were always constrained by the following rules of conditional prevalence obeyed within each realization:
\begin{equation}
\label{alphabet_hpair}
    p(f(t_{r+1,c}) = Y | f(t_{r,c}) = X) = 1,
\end{equation}
\begin{equation}
\label{alphabet_vpair}
    p(f(t_{r,c}) = Z | f(t_{r,c+1}) \in \{V, W, K\}) = 1.
\end{equation}
That is, the letter Y was always preceded horizontally by the letter X (Eq.\ref{alphabet_hpair}), and the letters V, W, K were always preceded vertically by the letter Z (Eq.\ref{alphabet_vpair}). Thus, four \textit{ordered} letter-pairs occured in each realization: X-Y (horizontal adjacency), and Z-K, Z-V, and Z-W (vertical adjacency). Furthermore, the per-realization prevalences of the letter-pairs were fixed as 8, 2, 1 and 1 respectively. For post-hoc processing, error for each $t$ was computed as the pixel-wise difference from the known letter templates and a reasonable acceptance threshold (75\% of the error scale maximum) was chosen once by visual inspection. Although the post-hoc classifier assigns an identity to all letters, only automatically recognizable letters were retained. This abates the effect of minor feature shape variance from further analysis. 

\subsection{Network trainings}
Two popular GAN architectures: ProGAN (PG) \cite{karras2017progressive} and StyleGAN2: config-e (SG) \cite{karras2020analyzing} were employed for this work. The prescribed default training schedule was found to be sufficient for training in terms of visual quality, Fr\'echet Inception Distance (FID) 10k scores \cite{heusel2017gans} and loss curve convergence. The trainings were performed such that the discriminator was shown 12 million images and 25 million images for PG and SG respectively; these were also the prescribed default training durations. For SG, the regularization parameter $R_1$ was set to 100 and the truncation parameter $\psi$ was set to 0.5; both are default values for the chosen configuration. The trainings were performed on Nvidia GeForce GTX 1080Ti, 1080, Tesla V100 and A100 GPUs, and typically took between 2 and 14 days per training on a single GPU. A total of 10240 realizations, for each dataset, were generated from each network for further analysis. It is explicitly noted that the goal of this work was not to achieve the best possible performance of any network, but simply to demonstrate the utility of the designed SCMs for assessing common DGMs that are trained in a typical way.

\section{Results}
\begin{figure}[htbp]
\centering
\begin{subfigure}{0.45\textwidth}%
    \includegraphics[width=0.9\linewidth]{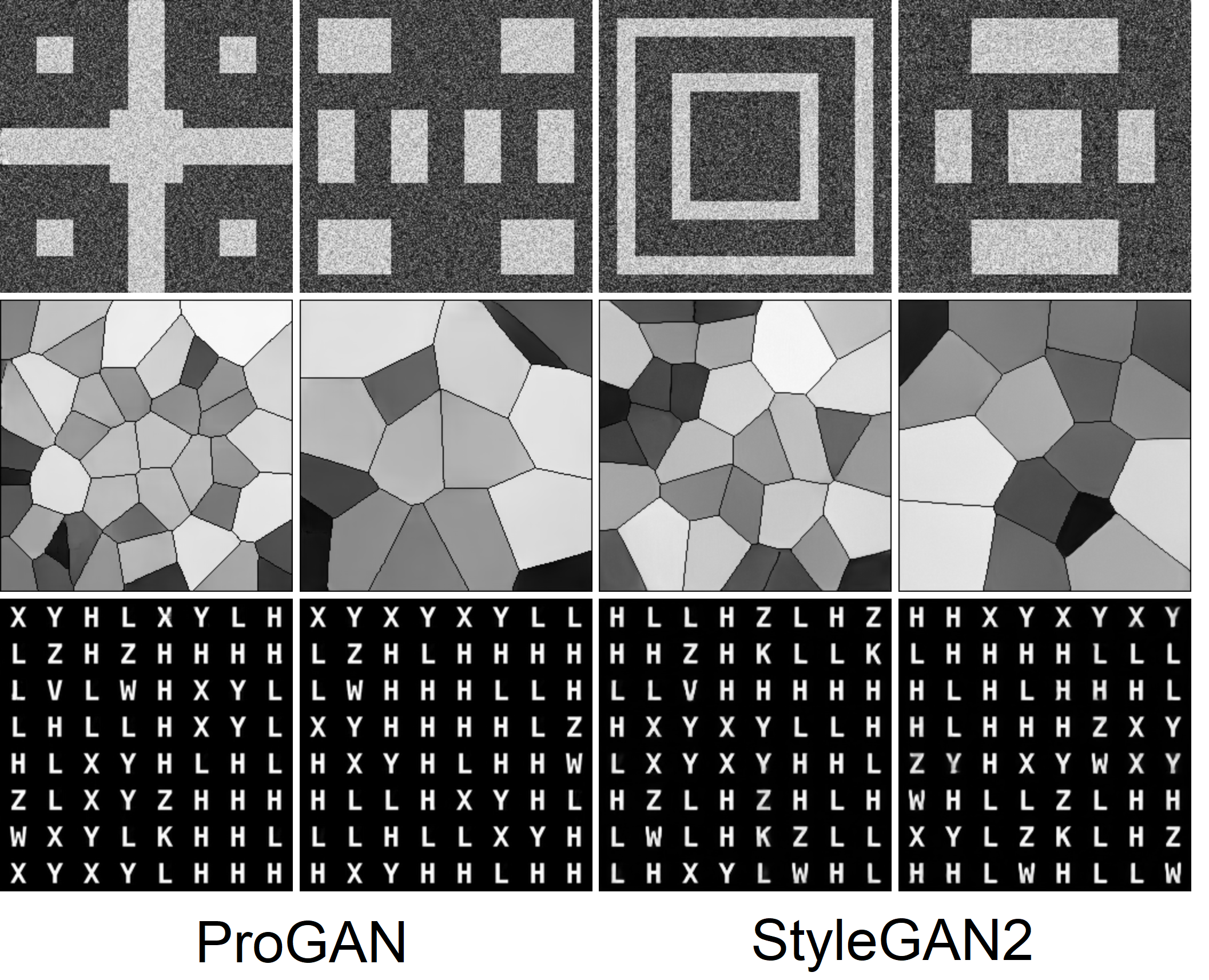} 
\end{subfigure}
\caption{Subjectively visually good GEN examples from networks trained on the three SCMs. Columns 1 and 2 show PG images while Columns 3 and 4 show SG images. Although the images demonstrate good visual similarity, contextual errors can be present in any image in any ensemble.}
\label{good_fakes}
\end{figure}

\begin{figure}[htbp]%
\centering
\begin{subfigure}{0.11\textwidth}%
\includegraphics[width=0.9\linewidth]{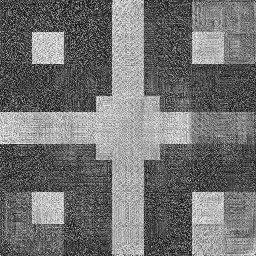} 
\end{subfigure}
\hspace{0.05em}
\begin{subfigure}{0.11\textwidth}%
\includegraphics[width=0.9\linewidth]{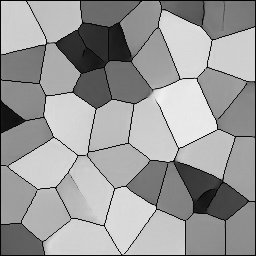} 
\end{subfigure}%
\hspace{0.05em}
\begin{subfigure}{0.11\textwidth}%
\includegraphics[width=0.9\linewidth]{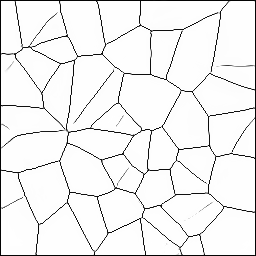} 
\end{subfigure}%
\hspace{0.05em}
\begin{subfigure}{0.11\textwidth}%
\includegraphics[width=0.9\linewidth]{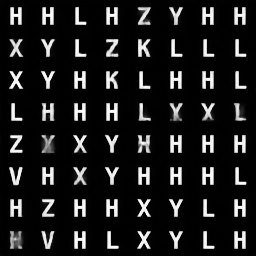} 
\end{subfigure}%
\caption{Class-mixing and artifacts in GEN images. GEN images occasionally exhibit artifacts such as blending of class-specific foregrounds in the flags SCM (left), weak boundaries and shading variance within distinct Voronoi regions (middle), and badly formed letters in the alphabet SCM (right).}
\label{artifacts}
\end{figure}

Sample generated images from both networks and all three SCMs are shown in Fig.~\ref{good_fakes} while examples of artifacts are shown in Fig.~\ref{artifacts}. The FID-10k scores \cite{heusel2017gans} for all models from both networks were between 2 and 10. Ensemble intensity distributions were also well replicated in all generated (GEN) ensembles.

\subsection{Results from the flags SCM}
Post-hoc processing of the GEN ensembles demonstrated that perfect match with the foreground templates was achieved for about 98\% realizations, while occasional malformations via blending of foreground templates was observed in the remaining cases. However, the forbiddance rule in Eq.\ref{flags_forbidden} was always respected. Realizations that did not perfectly match the original class templates were excluded from further analysis and several of those retained were visually spot-checked to ensure that they were well-formed. This abated the effect of foreground formation error and post-hoc processing on the statistical analysis of generated realizations. 

Equations \ref{flags_fg} and \ref{flags_bg}, representing intensity distribution requirements, were tested against a generous tolerance of 99.5th percentile of the chi-square statistic computed separately for \textit{f} and \textit{b}. None of the realizations generated from either network satisfied Eq.\ref{flags_fg} while about 1\% and 91\% images violated Eq.\ref{flags_bg} for PG and SG respectively, suggesting that the foreground and background were learned differently. This further indicates that first-order statistics computed from the foreground and background intensity distributions, could fail to match those of the training data. Such a failure not only implies that the distinct feature-specific foreground and background intensity distributions are not learned, but also that the application of a statistical observer or post-processing task such as thresholding or segmentation, could be adversely affected. Next, randomness in pixel placement, was tested via the tile-wise computation of Moran's I ($MI$) of spatial autocorrelation \cite{moran1950notes} for each $f_k$ and $b_k$ in every $I$. A tile was considered acceptable if the $MI$ was within $0\pm\sigma_M/256$, where $\sigma_M$ is the standard deviation of the distribution of the $MI$ computed on the training data, and a realization was considered acceptable if at most 3 tiles were rejected. On average, 3\% and 11\% of the realizations violated the distribution of $MI$ for the foreground and background for PG, while the proportion was about 4\% for both subsets for SG. These results imply that a majority of the realizations in ensembles generated from either network reproduce randomness in pixel arrangement. However, a non-negligible proportion, up to 1 in 9, of the realizations did not exhibit the prescribed randomness, and thus, inference based on the presumption of randomness could be incorrect. It was observed that the mean class prevalence matched the expected mean of 1/8, corresponding to uniform class prevalence in the training ensemble. Although the standard deviation was likely negligible for PG ($\sigma$=1\%), it was non-negligible for SG ($\sigma$=9\%), indicating that some classes were preferentially generated in the latter case. Thus, the relevant prevalence in a training data might not be reproduced in a GEN ensemble---this might have significant implications when employed for data augmentation or statistical power calculations. Thus, \emph{for this one SCM}, both second-order features and the per-image prevalence of second-order features was reasonably well reproduced; however, the first-order information per-image was essentially always wrong, even though the ensemble mean intensity distribution appears correct.

\subsection{Results from the Voronoi SCM}

\begin{figure}[h!tb]%
\centering
\begin{subfigure}{0.235\textwidth}%
\includegraphics[width=\linewidth]{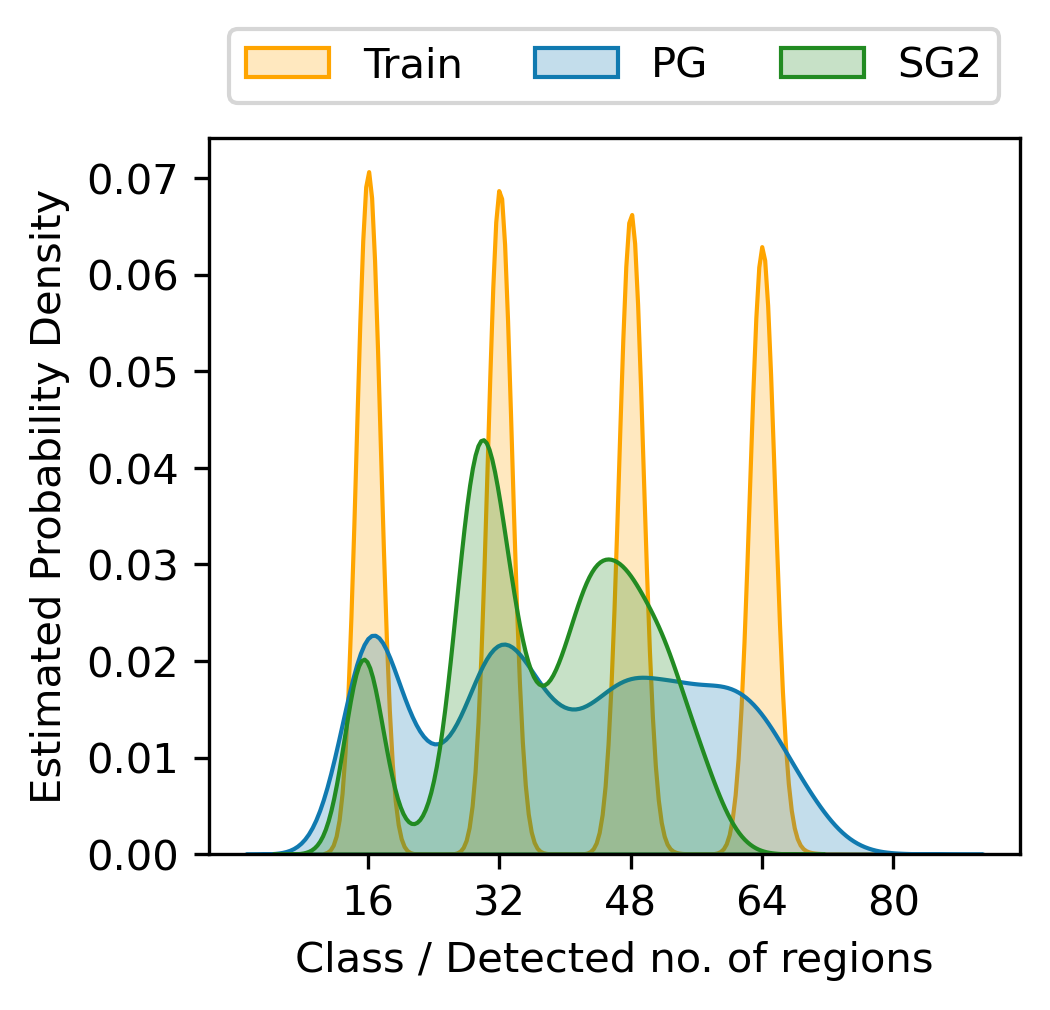} 
\end{subfigure}
\begin{subfigure}{0.235\textwidth}%
\includegraphics[width=\linewidth]{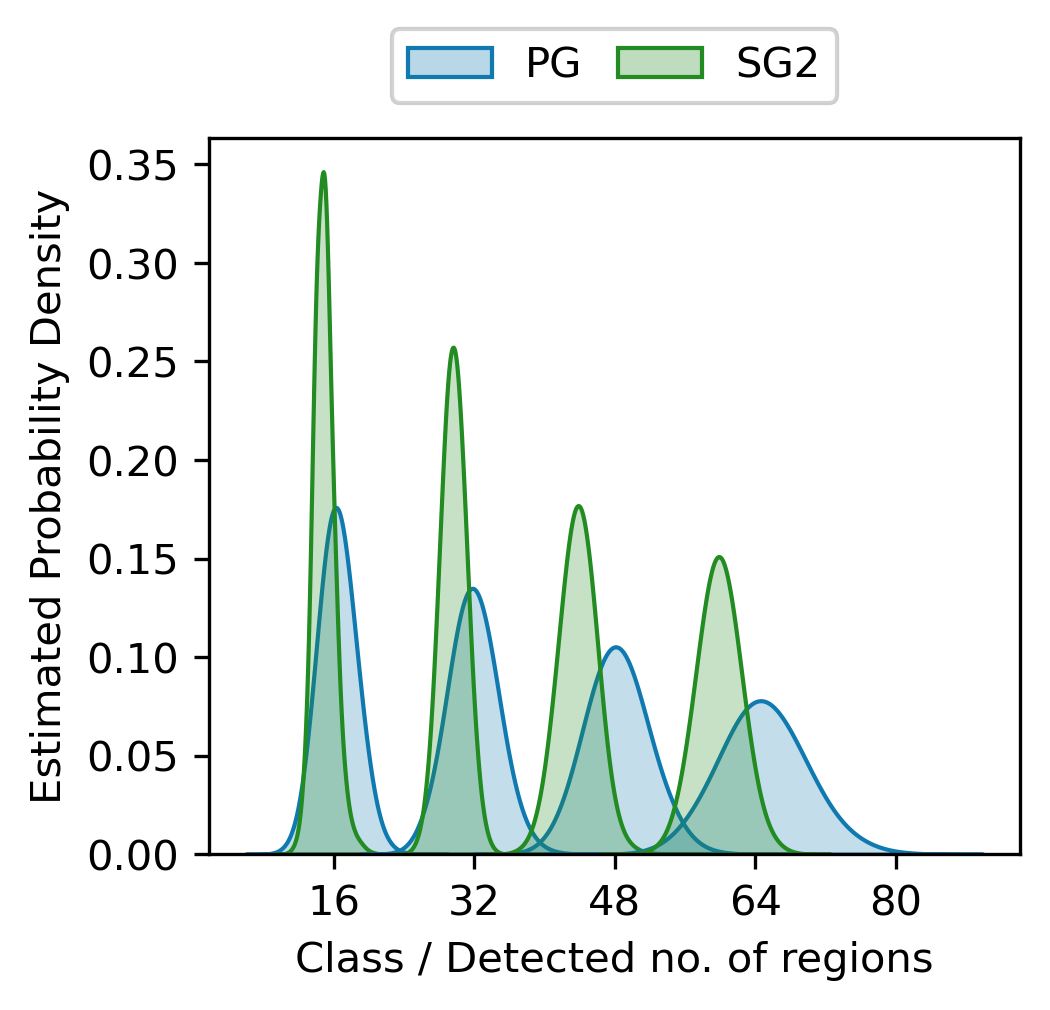} 
\end{subfigure}%
\caption{Results from the Voronoi SCM for class prevalence studies. Left: The expected equal class prevalence in the ensemble was not reproduced in the GEN images from both networks, but more significantly for SG. The effect of error from the post-hoc classifier is also observed. Right: Four separate models, each trained on a single class, generated images outside the class for both networks. While the SG-generated ensemble demonstrated class extrapolation, the PG-generated ensemble showed slightly shifted class means.}
\label{voronoi_prevalence}
\end{figure}

\begin{figure}[htbp]%
\centering
\begin{subfigure}{0.23\textwidth}%
\includegraphics[width=\linewidth]{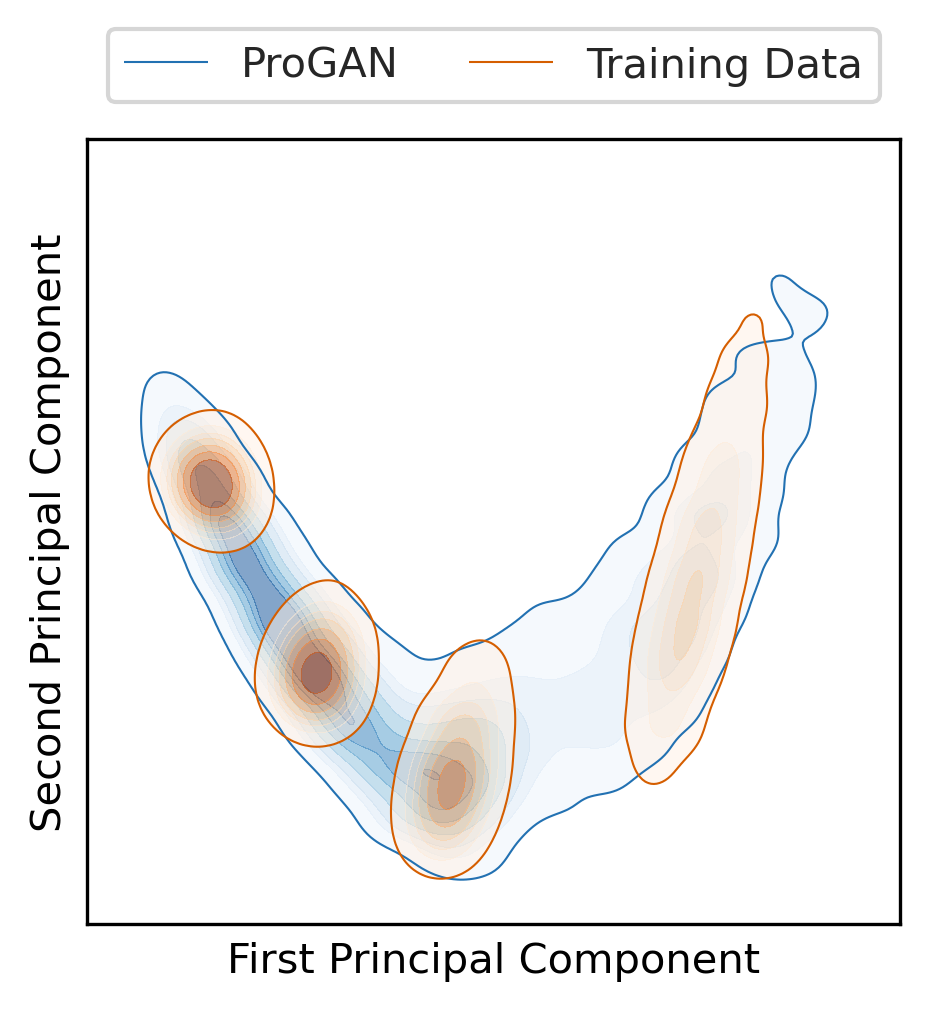} 
\end{subfigure}
\begin{subfigure}{0.23\textwidth}%
\includegraphics[width=\linewidth]{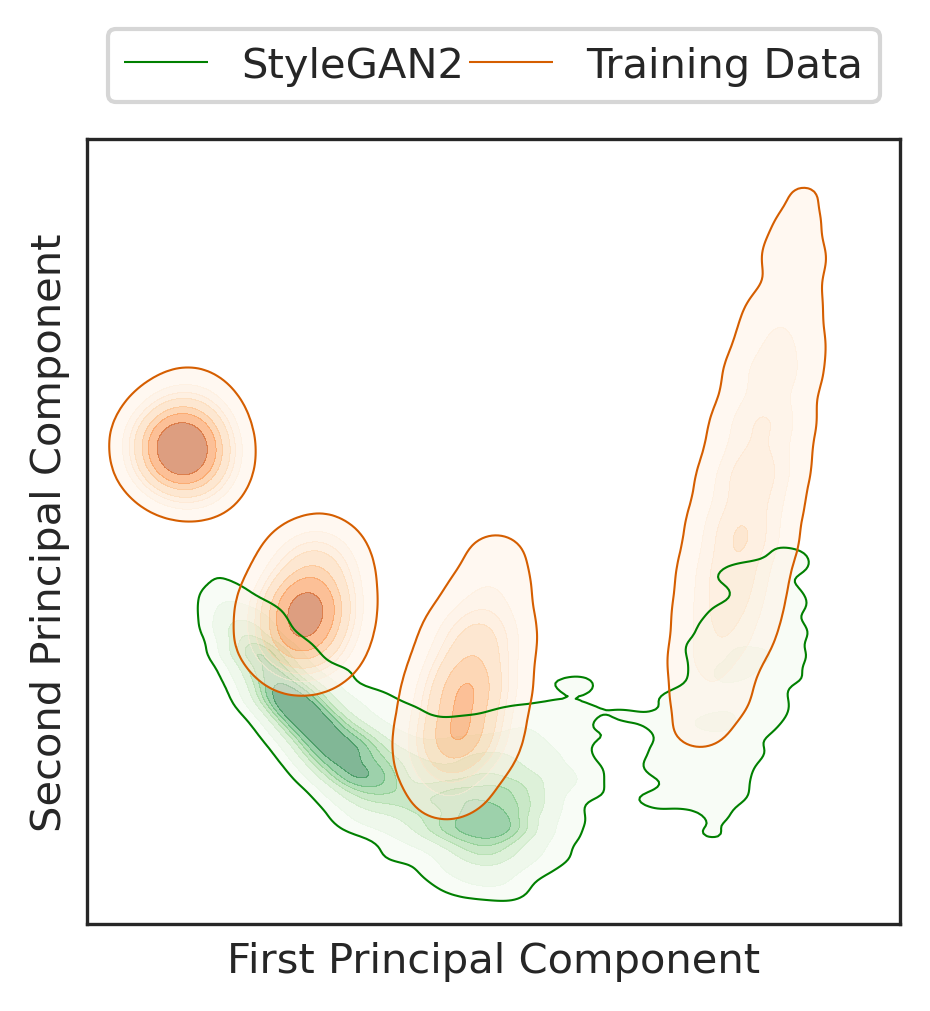} 
\end{subfigure}%
\caption{Results from the Voronoi SCM for assessment of implicit context. Statistics representing implicit context were projected onto the two highest principal components for true and PG-GEN (left) and SG-GEN (right) ensembles. Interpolation between the four training classes was more prevalent in the PG-GEN ensemble while lower overlap in feature clouds was observed in the SG-GEN ensemble, indicating dissimilar ranges of these statistics in the latter.}
\label{pca_plots}
\end{figure}

Although high visual similarity was observed in the GEN Voronoi images, various artifacts were also observed such as: the presence of (i) low-amplitude artifacts in regions of constant intensity \cite{deshpande2022evaluating}, (ii) curved or floating region edges and (iii) multiple intensity values instead of only one in a single Voronoi region (see Fig.~\ref{artifacts}). Low-amplitude artifacts, possibly characteristic of the convolutional network architecture, could affect decision-making because the original second-order information---and thus, possibly any derived texture statistic---is not consistent with the original dataset. The other artifacts, visually more apparent, can confound any classifier or analysis that is calibrated on the training data. Furthermore, the high-order rule in Eq.\ref{voronoi} relating intensity and area of a shaded Voronoi region, was tested and it was observed that the rule was not reproduced exactly. The expected Spearman rank correlation ($\rho$=1.0) was lowered in the GEN images. A decrease of over 20\% ($\rho<0.8$) was observed in 3\% and 2\% realizations from PG and SG respectively. If the grayscale intensity $g$ represents a physical property, violation of Eq.\ref{voronoi} implies that these realizations have at least partially lost their quantitative meaning. Next, studies of class prevalence were performed with the Voronoi SCM by training five different models for each network architecture on the training data representing: (i) all four classes equally, and (ii-v) each of the four classes individually. As seen in Fig.~\ref{voronoi_prevalence}, class prevalence in case (i) was not maintained in the ensemble generated from either network whereas class extrapolation was observed in cases (ii-v). 

\begin{figure}[h!tb]
\centering
\begin{subfigure}{0.46\textwidth}%
    \includegraphics[width=\linewidth]{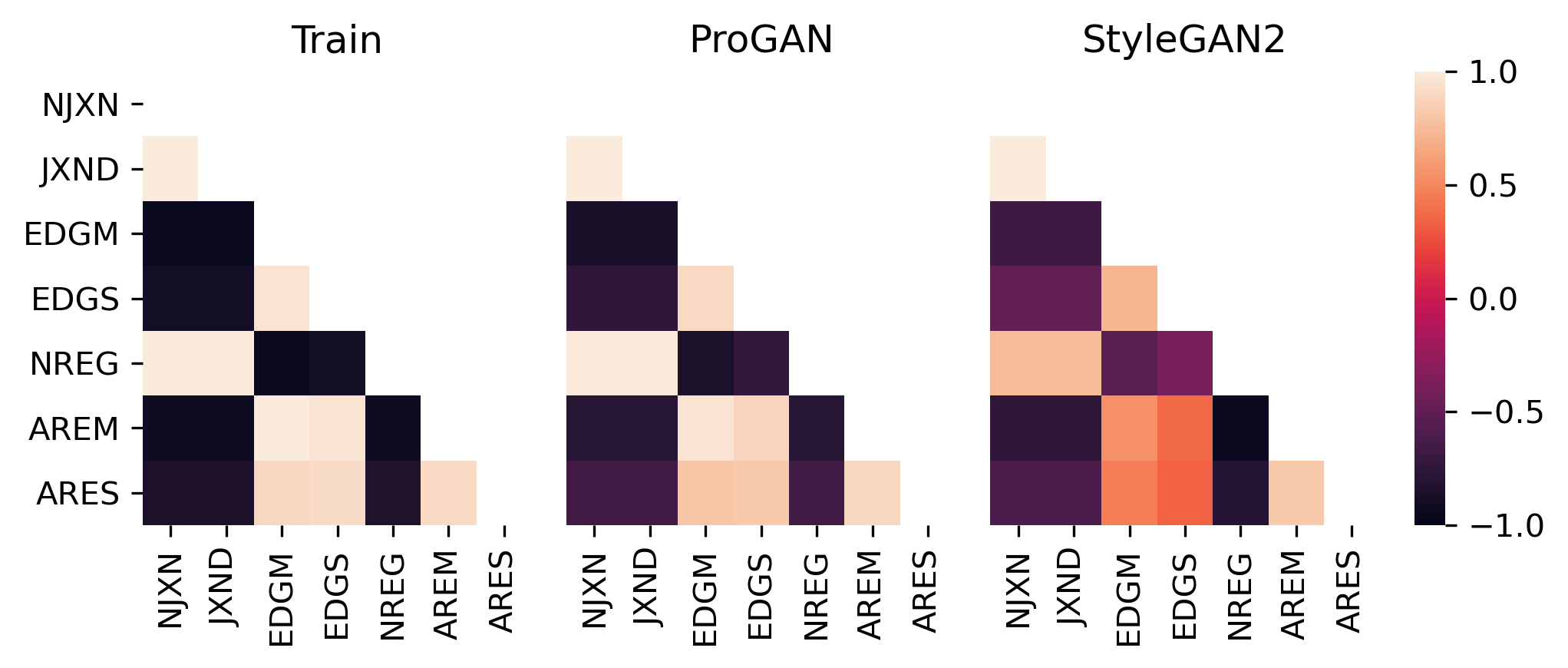} 
\end{subfigure}
\caption{Results from the \emph{unshaded} Voronoi SCM for assessment of implicit context. The strengths of correlations of the per-image statistics representing implicit context (left) were lowered in both GEN ensembles (center and right), especially in SG, indicating that the correct implicit context was not reproduced.}
\label{heatmaps}
\end{figure}

Last, implicit context in Voronoi diagrams was assessed under case (i) by employing the Skan Python library \cite{nunez2018new} to compute the following statistics derived from Voronoi regions and edges: number of junctions (NJXN), junction density (JXND), mean edge length (EDGM), standard deviation over edge lengths (EDGS), number of regions (NREG), mean area of a region (AREM), and standard deviation over region area (ARES). Interpolation between class-specific features was observed (see Fig.~\ref{pca_plots}) via principal component analysis (PCA) of the features listed above. This indicates extrapolation in the learned feature space. Even when the classes (or NREG) were incorrect due to extrapolation, the implicit context was generally retained in this case. Reproduction of implicit context was then tested in the absence of shading, that is, by employing the \emph{unshaded} Voronoi SCM. Partial loss of implicit context was observed via decreased correlations between the studied statistics (see Fig.~\ref{heatmaps}) and lower overlap in the feature clouds in PCA (not shown) as compared to Fig.~\ref{pca_plots}. Further confirmation of implicit contextual errors in the GEN ensemble was obtained via testing two well established statistical properties of Voronoi diagrams. Specifically, Property V11-1 and V12 (here onward referred to as P1 and P2) in \cite{boots2009spatial} were tested. Both conditions were satisfied in 99\% of the training data. For the unshaded Voronoi, the rates of violation for both conditions in PG-GEN and SG-GEN ensembles respectively were: 12\% and 100\% for P1, and 10\% and 100\% for P2. For the shaded Voronoi, these rates were under 6\% for both GEN ensembles. Note that it is possible that a different model or training strategy may have fewer implicit contextual errors. Here, we only demonstrate that implicit contextual errors made by a model trained in a typical manner and achieving low FID-10k scores, can be detected via the proposed method. These results suggest that the reproduction of implicit context even for datasets such as the unshaded Voronoi is a non-trivial task for a GAN and may have significant implications in domain-specific space partitioning problems. 

\subsection{Results from the alphabet SCM}
Although most letters were well-formed in the GEN ensembles, errors occasionally were observed as seen in Fig.~\ref{artifacts}. Error rates of letter formation via post-hoc processing were: 1 in 6250 letters for PG and 1 in 73 letters for SG respectively, indicating that almost all letters in a realization were recognizable. Although relatively few letters were unrecognizable, only images where all letters were recognizable, 99\% and 59\% of the ensemble for PG and SG respectively, were considered for further analysis. A sample of 10000 such well-formed realizations was explicitly tested for high-order rules of feature prevalence.  

\begin{figure}[htbp]
\centering
\includegraphics[width=0.7\linewidth]{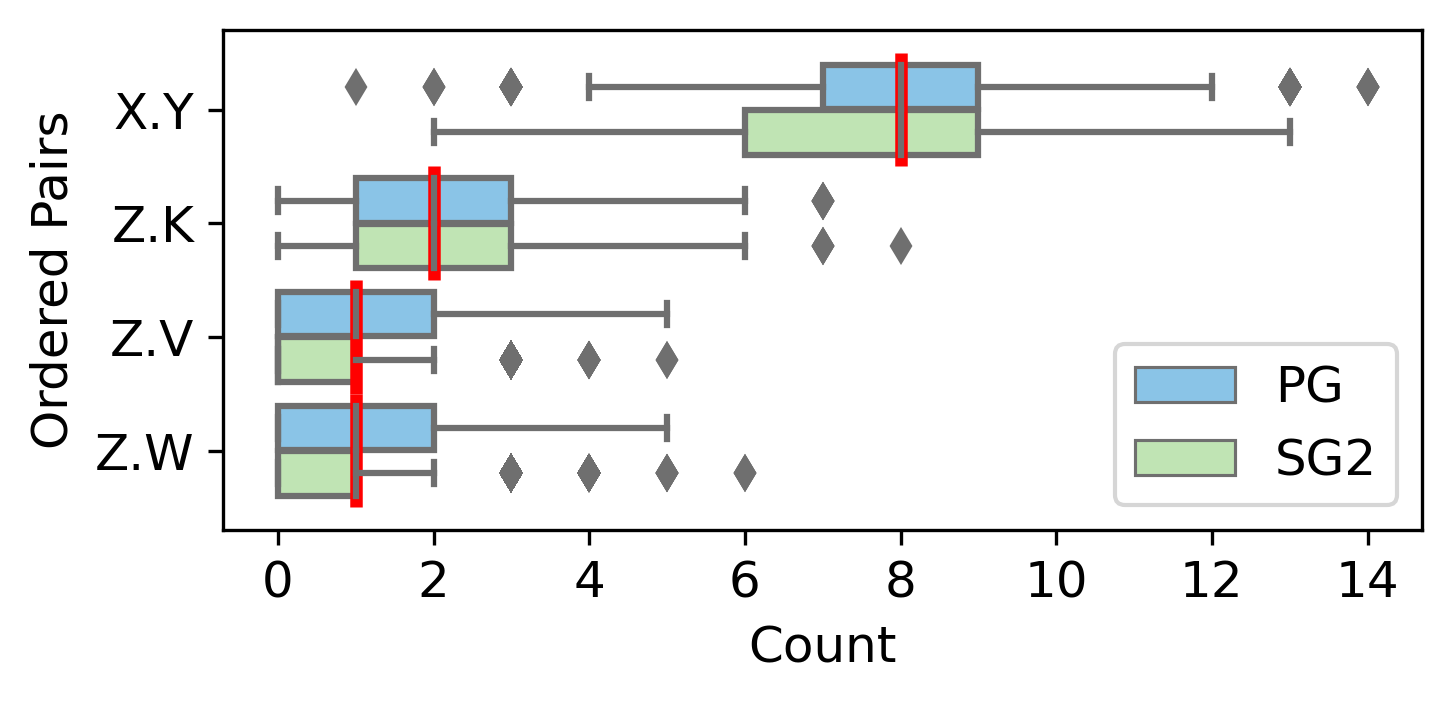} 
\caption{Results from the alphabet SCM. The expected paired-letter prevalences: X-Y, Z-K, Z-V, Z-W = 8, 2, 1, 1 were not respected by either network. Correct prevalence is marked in red. A wide range of values was seen for both networks indicating that perfect prevalence is achieved only by chance.}
\label{alphabet_boxplots}
\end{figure}

The observed frequency of letters was compared to the prescribed set $\mathbb{B}$ via the $\chi^2$ goodness-of-fit test. Only 119 PG and 72 SG realizations were found to be outside the 95\% critical value of the chi-squared test. However, this means only that the letters appear to have been drawn from the prescribed distribution, not that a realization is correct. In fact, on testing per-image letter prevalences (Eq.\ref{alphabet_main}), we observed that only 18 PG realizations and 6 SG realizations exactly matched $\mathbb{B}$; recall, this frequency is identical in \emph{every} training image. Thus, only by rare chance was any realization correct in high order. Incidentally, we observed that these were not memorized realizations. In a certain domain, if natural variation exists in the prevalence of a feature, most realizations would be acceptable. However, if the feature prevalence is the context required for a downstream task, then essentially none of the realizations from these GEN ensembles are acceptable. Next, the prescribed ordered pair-prevalences (Eqs.8 and 9) were tested. The fixed ordered pairs X-Y, Z-K, Z-V and Z-W were expected to occur at frequencies of precisely 8, 2, 1 and 1 respectively, but were observed to occur at a wide range of frequencies (see Fig. \ref{alphabet_boxplots}) for both networks. The single letters V, W and K which \underline{never} occur without the partner in the training data, occurred without the other member of the pair up to 100\% of time, similarly the letter Y occurred separately about 37\% of time for PG. This rate of separate occurrence of letters in letter-pairs is approximately doubled or tripled for SG. Hence, pairs of image features that may be expected to have known, relative locations and prevalence might not appear in a GEN ensemble. Thus, ``visually good'' GEN images might have diminished domain-specific value due to an unrealistic representation. 

\section{Discussion}
Much improvement has occurred in the realism of GEN natural images and their evaluation \cite{shamsolmoali2021image, borji2021pros}. However, the deployment of GANs in domains where domain expertise is inextricably tied to image perception, such as in medical imaging, still remains a challenge \cite{astley2020special}. To partially circumvent this challenge, the proposed SCMs provide a method for encoding high-order information relevant to a domain while also allowing the recovery of this information from a GEN ensemble. In the present work, we represented high-order information via explicit modeling of contexts such as feature prevalences and relative feature arrangements, but the use of SCMs is not limited to these scenarios. Any other representation of spatial context that may be relevant to a certain domain could be employed similarly for evaluation as long its recovery from the generated ensemble is sufficiently robust. The SCM-based method of GAN evaluation can also be further developed for a specific domain as demonstrated in \cite{kelkar2023assessing} for medical imaging.

Although some works have studied the reproducibility of long distance spatial context by generative models \cite{pmlr-v139-nash21a, tsitsulin2019shape}, these methods do not employ purposefully created datasets for evaluation. As the proposed method of evaluation is data-centric and independent of the generative model type, it can be readily employed on any generative model: deep or conventional. Thus, it may enable the benchmarking of new architectures against existing architectures or aid the design and development process of generative models for domain-specific applications. 

Of course, each instance of a particular GAN is unique, and thus the results may vary between instances, however, any instances trained from the same architecture share some common learning capacity. We envision the use of designed SCMs as a kind of ``necessary but not sufficient'' triage of GAN capacity. Our supposition is that if a particular GAN demonstrably fails at recovering the fundamental image properties one prescribes---such as grayscale intensity distribution, spatial randomness, and pre-specified feature prevalences ---then that architecture could fail to accurately reproduce any sort of domain-relevant image that comprises those fundamental properties. This is why SCMs such as the ones we have designed can be relevant to estimating the probability that a GAN has made errors in domain-specific images. In future, we intend to extend the method of evaluating GANs via SCMs to tasks other than unconditional synthesis, such as conditional synthesis and de-noising. 

\subsection{Interpretation of results within a specific domain: diagnostic imaging}
In medical images, features can have quantitative, structural, and positional significance within each realization; this can be partially described by statistics spanning multiple orders of information. However, the joint reproduction of statistics across multiple orders might be a challenging task for the chosen GANs as observed in the results from the flags SCM and, hence, the ultimate utility of a generated realization might be determined by the order of information required for a specific diagnostic task. For example, a GAN employed for simulating positron emission tomography images of a certain tumor type may produce a majority tumors of correct shape but significantly different in the expected intensity distribution and texture. Drawing diagnostic inferences from such a generated ensemble, even when employed for data augmentation, might translate to false clinical predictions.

The Voronoi SCM was designed such that image features, corresponding to the Voronoi regions, were ergodic. An analogous clinical example is a histology image, depicting multiple cell types, each with characteristic textural features and staining intensity but able to appear anywhere in the field of view. When the rank-correlation between area and grayscale intensity was not reproduced correctly, the quantitative information---here, representative of physical tissue properties---could be unreliable and the derived textural features suspect. Furthermore, for both multi-class SCMs: flags and Voronoi, the incorrectly reproduced class prevalence in the generated ensembles suggests that if these particular instances of GANs were used to replicate a clinical dataset for virtual clinical trials, or to generate a training ensemble for a downstream task, the prevalence of the input pathologies would not be maintained. Most significantly, this bias could be characteristic of the network-architecture and thus, would have to be quantified for each architecture separately.

The alphabet SCM was designed with known per-realization prevalence of single and paired features in order to isolate reproducibility of high-order features (the letters) from that effects of variable position, structure and shading. Because anatomical plausibility can be represented (at least partially) as the joint, per-realization prevalence of naturally occurring features, it is paramount that this prevalence is maintained within each realization and not just on average, over the ensemble. If a generative model is designed to maximize similarity over the ensemble alone, per-realization errors might be widespread as was observed in the GEN images of the alphabet SCM where less than 0.2\% of the ensemble had perfect feature and feature-pair prevalence. Such visually realistic GEN images with incorrect per-realization feature prevalence might have reduced diagnostic value. This could translate into a bias in, or even a complete failure to learn, the information required for particular decision tasks.

\section{Conclusion}
The main conclusion is that SCMs can be designed to enable the quantification of certain impactful, per-realization errors made by some popular GAN architectures at a high rate even when summary and ensemble measures of training appear reasonable. The main reason that these errors are difficult to evaluate in scenarios requiring a substantial domain expertise is that there usually is not a mathematically specified ground truth or expert labeling for each generated realization. 

In this work, it is demonstrated how stochastic context models can be purposefully designed to include known high-order
contextual information, analogous to domain-relevant external information, that also can be quantified post-generation and thus serve as a ground truth. This design can be done algorithmically, without actually specifying a formula for any particular high-order statistic. Several such SCMs were proposed and employed in the evaluation of two popular GAN architectures. 

Across various training and model scenarios, it was found that the tested models failed to simultaneously reproduce all prescribed contextual features, at once, despite being well replicated in the ensemble, and despite obvious visual similarity between training and generated data. Specifically, numerous per-realization errors occur in: grayscale intensity distribution, spatial arrangement of those intensities, and, perhaps most impactful, in the frequency of pre-specified rates of feature occurrence. Here, we do not claim that one architecture is better than the other, but that observable differences between the chosen instances of the two architectures could be exposed by the use of the proposed method.

The corollary is that the designed SCMs can serve as a kind of triage before even more sophisticated task-based measures of generated image quality are employed, or as benchmarking datasets for advancing generative model design.

\section{Acknowledgements}
This work was supported in part by NIH awards EB020604, EB023045 and EB031772. RD acknowledges support from the WU-ISP Trainee Fellowship (5T32 EB01485505). This work utilized resources supported by the NSF grants 1725729, OCI 2005572, NCSA, UIUC and the State of Illinois.

\bibliographystyle{model1-num-names}
\bibliography{references}

\begin{thebibliography}{18}
\expandafter\ifx\csname natexlab\endcsname\relax\def\natexlab#1{#1}\fi
\providecommand{\url}[1]{\texttt{#1}}
\providecommand{\href}[2]{#2}
\providecommand{\path}[1]{#1}
\providecommand{\DOIprefix}{doi:}
\providecommand{\ArXivprefix}{arXiv:}
\providecommand{\URLprefix}{URL: }
\providecommand{\Pubmedprefix}{pmid:}
\providecommand{\doi}[1]{\href{http://dx.doi.org/#1}{\path{#1}}}
\providecommand{\Pubmed}[1]{\href{pmid:#1}{\path{#1}}}
\providecommand{\bibinfo}[2]{#2}
\ifx\xfnm\relax \def\xfnm[#1]{\unskip,\space#1}\fi
\bibitem[{Bond-Taylor et~al.(2021)Bond-Taylor, Leach, Long, and
  Willcocks}]{bond2021deep}
\bibinfo{author}{S.~Bond-Taylor}, \bibinfo{author}{A.~Leach},
  \bibinfo{author}{Y.~Long}, \bibinfo{author}{C.~G. Willcocks},
\newblock \bibinfo{title}{Deep generative modelling: A comparative review of
  {VAE}s, {GAN}s, normalizing flows, energy-based and autoregressive models},
\newblock \bibinfo{journal}{IEEE Trans. Pattern Anal. Mach. Intell.}
  (\bibinfo{year}{2021}).
\bibitem[{Goodfellow et~al.(2020)Goodfellow, Pouget-Abadie, Mirza, Xu,
  Warde-Farley, Ozair, Courville, and Bengio}]{goodfellow2020generative}
\bibinfo{author}{I.~Goodfellow}, \bibinfo{author}{J.~Pouget-Abadie},
  \bibinfo{author}{M.~Mirza}, \bibinfo{author}{B.~Xu},
  \bibinfo{author}{D.~Warde-Farley}, \bibinfo{author}{S.~Ozair},
  \bibinfo{author}{A.~Courville}, \bibinfo{author}{Y.~Bengio},
\newblock \bibinfo{title}{Generative adversarial networks},
\newblock \bibinfo{journal}{Commun. ACM} \bibinfo{volume}{63}
  (\bibinfo{year}{2020}) \bibinfo{pages}{139--144}.
\bibitem[{Kazeminia et~al.(2020)Kazeminia, Baur, Kuijper, van Ginneken, Navab,
  Albarqouni, and Mukhopadhyay}]{kazeminia2020gans}
\bibinfo{author}{S.~Kazeminia}, \bibinfo{author}{C.~Baur},
  \bibinfo{author}{A.~Kuijper}, \bibinfo{author}{B.~van Ginneken},
  \bibinfo{author}{N.~Navab}, \bibinfo{author}{S.~Albarqouni},
  \bibinfo{author}{A.~Mukhopadhyay},
\newblock \bibinfo{title}{{GANs} for medical image analysis},
\newblock \bibinfo{journal}{Artif. Intell. in Med.} \bibinfo{volume}{109}
  (\bibinfo{year}{2020}) \bibinfo{pages}{101938}.
\bibitem[{Shamsolmoali et~al.(2021)Shamsolmoali, Zareapoor, Granger, Zhou,
  Wang, Celebi, and Yang}]{shamsolmoali2021image}
\bibinfo{author}{P.~Shamsolmoali}, \bibinfo{author}{M.~Zareapoor},
  \bibinfo{author}{E.~Granger}, \bibinfo{author}{H.~Zhou},
  \bibinfo{author}{R.~Wang}, \bibinfo{author}{M.~E. Celebi},
  \bibinfo{author}{J.~Yang},
\newblock \bibinfo{title}{Image synthesis with adversarial networks: A
  comprehensive survey and case studies},
\newblock \bibinfo{journal}{Inf. Fusion}  (\bibinfo{year}{2021}).
\bibitem[{Borji(2022)}]{borji2021pros}
\bibinfo{author}{A.~Borji},
\newblock \bibinfo{title}{Pros and cons of {GAN} evaluation measures: New
  developments},
\newblock \bibinfo{journal}{Comput. Vis. Image Underst.}
  (\bibinfo{year}{2022}).
\bibitem[{Barrett and Myers(2013)}]{barrett2013foundations}
\bibinfo{author}{H.~H. Barrett}, \bibinfo{author}{K.~J. Myers},
  \bibinfo{title}{Foundations of image science}, \bibinfo{publisher}{John Wiley
  \& Sons}, \bibinfo{year}{2013}.
\bibitem[{Badano(2017)}]{badano2017much}
\bibinfo{author}{A.~Badano},
\newblock \bibinfo{title}{“{How} much realism is needed?”—the wrong
  question in silico imagers have been asking},
\newblock \bibinfo{journal}{Med. Phys.} \bibinfo{volume}{44}
  (\bibinfo{year}{2017}) \bibinfo{pages}{1607--1609}.
\bibitem[{Deshpande et~al.(2022)Deshpande, Anastasio, and
  Brooks}]{deshpande2022evaluating}
\bibinfo{author}{R.~Deshpande}, \bibinfo{author}{M.~A. Anastasio},
  \bibinfo{author}{F.~J. Brooks},
\newblock \bibinfo{title}{Evaluating the capacity of deep generative models to
  reproduce measurable high-order spatial arrangements in diagnostic images},
\newblock in: \bibinfo{booktitle}{SPIE Medical Imaging}, volume
  \bibinfo{volume}{12032}, \bibinfo{organization}{SPIE}, \bibinfo{year}{2022},
  pp. \bibinfo{pages}{521--526}.
\bibitem[{Boots et~al.(2009)Boots, Sugihara, Chiu, and
  Okabe}]{boots2009spatial}
\bibinfo{author}{B.~Boots}, \bibinfo{author}{K.~Sugihara},
  \bibinfo{author}{S.~N. Chiu}, \bibinfo{author}{A.~Okabe},
  \bibinfo{title}{Spatial tessellations: concepts and applications of Voronoi
  diagrams}, \bibinfo{publisher}{John Wiley \& Sons}, \bibinfo{year}{2009}.
\bibitem[{Karras et~al.(2018)Karras, Aila, Laine, and
  Lehtinen}]{karras2017progressive}
\bibinfo{author}{T.~Karras}, \bibinfo{author}{T.~Aila},
  \bibinfo{author}{S.~Laine}, \bibinfo{author}{J.~Lehtinen},
\newblock \bibinfo{title}{Progressive growing of {GAN}s for improved quality,
  stability, and variation},
\newblock \bibinfo{journal}{International Conference on Learning
  Representations}  (\bibinfo{year}{2018}).
\bibitem[{Karras et~al.(2020)Karras, Laine, Aittala, Hellsten, Lehtinen, and
  Aila}]{karras2020analyzing}
\bibinfo{author}{T.~Karras}, \bibinfo{author}{S.~Laine},
  \bibinfo{author}{M.~Aittala}, \bibinfo{author}{J.~Hellsten},
  \bibinfo{author}{J.~Lehtinen}, \bibinfo{author}{T.~Aila},
\newblock \bibinfo{title}{Analyzing and improving the image quality of
  style{GAN}},
\newblock in: \bibinfo{booktitle}{IEEE Conference on Computer Vision and
  Pattern Recognition}, \bibinfo{year}{2020}, pp. \bibinfo{pages}{8110--8119}.
\bibitem[{Heusel et~al.(2017)Heusel, Ramsauer, Unterthiner, Nessler, and
  Hochreiter}]{heusel2017gans}
\bibinfo{author}{M.~Heusel}, \bibinfo{author}{H.~Ramsauer},
  \bibinfo{author}{T.~Unterthiner}, \bibinfo{author}{B.~Nessler},
  \bibinfo{author}{S.~Hochreiter},
\newblock \bibinfo{title}{Gans trained by a two time-scale update rule converge
  to a local nash equilibrium},
\newblock \bibinfo{journal}{Advances in Neural Information Processing Systems}
  \bibinfo{volume}{30} (\bibinfo{year}{2017}).
\bibitem[{Moran(1950)}]{moran1950notes}
\bibinfo{author}{P.~A. Moran},
\newblock \bibinfo{title}{Notes on continuous stochastic phenomena},
\newblock \bibinfo{journal}{Biometrika} \bibinfo{volume}{37}
  (\bibinfo{year}{1950}) \bibinfo{pages}{17--23}.
\bibitem[{Nunez-Iglesias et~al.(2018)Nunez-Iglesias, Blanch, Looker, Dixon, and
  Tilley}]{nunez2018new}
\bibinfo{author}{J.~Nunez-Iglesias}, \bibinfo{author}{A.~J. Blanch},
  \bibinfo{author}{O.~Looker}, \bibinfo{author}{M.~W. Dixon},
  \bibinfo{author}{L.~Tilley},
\newblock \bibinfo{title}{A new python library to analyse skeleton images
  confirms malaria parasite remodelling of the red blood cell membrane
  skeleton},
\newblock \bibinfo{journal}{PeerJ} \bibinfo{volume}{6} (\bibinfo{year}{2018})
  \bibinfo{pages}{e4312}.
\bibitem[{Astley et~al.(2020)Astley, Chen, Myers, and
  Nishikawa}]{astley2020special}
\bibinfo{author}{S.~M. Astley}, \bibinfo{author}{W.~Chen},
  \bibinfo{author}{K.~J. Myers}, \bibinfo{author}{R.~M. Nishikawa},
\newblock \bibinfo{title}{Special section guest editorial: Evaluation
  methodologies for clinical {AI}},
\newblock \bibinfo{journal}{J. Med. Imaging} \bibinfo{volume}{7}
  (\bibinfo{year}{2020}).
\bibitem[{Kelkar et~al.(2023)Kelkar, Gotsis, Brooks, Prabhat, Myers, Zeng, and
  Anastasio}]{kelkar2023assessing}
\bibinfo{author}{V.~A. Kelkar}, \bibinfo{author}{D.~S. Gotsis},
  \bibinfo{author}{F.~J. Brooks}, \bibinfo{author}{K.~Prabhat},
  \bibinfo{author}{K.~J. Myers}, \bibinfo{author}{R.~Zeng},
  \bibinfo{author}{M.~A. Anastasio},
\newblock \bibinfo{title}{Assessing the ability of generative adversarial
  networks to learn canonical medical image statistics},
\newblock \bibinfo{journal}{IEEE Trans. Med. Imaging}  (\bibinfo{year}{2023}).
\bibitem[{Nash et~al.(2021)Nash, Menick, Dieleman, and
  Battaglia}]{pmlr-v139-nash21a}
\bibinfo{author}{C.~Nash}, \bibinfo{author}{J.~Menick},
  \bibinfo{author}{S.~Dieleman}, \bibinfo{author}{P.~Battaglia},
\newblock \bibinfo{title}{Generating images with sparse representations},
\newblock in: \bibinfo{booktitle}{International Conference on Machine
  Learning}, volume \bibinfo{volume}{139}, \bibinfo{publisher}{PMLR},
  \bibinfo{year}{2021}, pp. \bibinfo{pages}{7958--7968}.
\bibitem[{Tsitsulin et~al.(2020)Tsitsulin, Munkhoeva, Mottin, Karras,
  Bronstein, Oseledets, and M{\"u}ller}]{tsitsulin2019shape}
\bibinfo{author}{A.~Tsitsulin}, \bibinfo{author}{M.~Munkhoeva},
  \bibinfo{author}{D.~Mottin}, \bibinfo{author}{P.~Karras},
  \bibinfo{author}{A.~Bronstein}, \bibinfo{author}{I.~Oseledets},
  \bibinfo{author}{E.~M{\"u}ller},
\newblock \bibinfo{title}{The shape of data: Intrinsic distance for data
  distributions},
\newblock in: \bibinfo{booktitle}{International Conference on Learning
  Representations}, \bibinfo{year}{2020}.

\end{thebibliography}

\end{document}